\documentclass[runningheads]{llncs}
\usepackage{graphicx}
\usepackage{comment}
\usepackage{amsmath,amssymb} 
\usepackage{color}

\usepackage{diagbox}
\usepackage{multirow}
\usepackage{array}
\usepackage{wrapfig}
\usepackage[misc]{ifsym} 

\begin{document}
	\pagestyle{headings}
	\mainmatter
	\def\ECCVSubNumber{3119}  
	
	\title{Learning From Multiple Experts: Self-paced Knowledge Distillation for Long-tailed Classification} 

	\titlerunning{LFME: Self-paced Knowledge Distillation for Long-tailed Classification}
	%
	\author{Liuyu Xiang \inst{1}  \and
		Guiguang Ding \inst{1,(\textrm{\Letter})}   \and
		Jungong Han \inst{2}}
	\authorrunning{L.Y. Xiang et al.}
	%
	\institute{School of Software, Tsinghua University, Beijing, China;\\
		Beijing National Research Center for Information Science and Technology (BNRist)
		\email{xiangly17@mails.tsinghua.edu.cn}, \email{dinggg@tsinghua.edu.cn}
		\and
		Computer Science Department, Aberystwyth University, SY23 3FL, UK \\
		\email{jungonghan77@gmail.com}}
	\maketitle


	\begin{abstract}
		In real-world scenarios, data tends to exhibit a long-tailed distribution, which increases the difficulty of training deep networks. In this paper, we propose a novel self-paced knowledge distillation framework, termed Learning From Multiple Experts (LFME). Our method is inspired by the observation that networks trained on less imbalanced subsets of the distribution often yield better performances than their jointly-trained counterparts. We refer to these models as `Experts’, and the proposed LFME framework aggregates the knowledge from multiple `Experts' to learn a unified student model. Specifically, the proposed framework involves two levels of adaptive learning schedules: Self-paced Expert Selection and Curriculum Instance Selection, so that the knowledge is adaptively transferred to the `Student'. We conduct extensive experiments and demonstrate that our method is able to achieve superior performances compared to state-of-the-art methods. We also show that our method can be easily plugged into state-of-the-art long-tailed classification algorithms for further improvements.
	\end{abstract}
	
	\section{Introduction}
	
	\begin{figure}[htbp!]
		\centering
		\includegraphics[width= 0.9\linewidth]{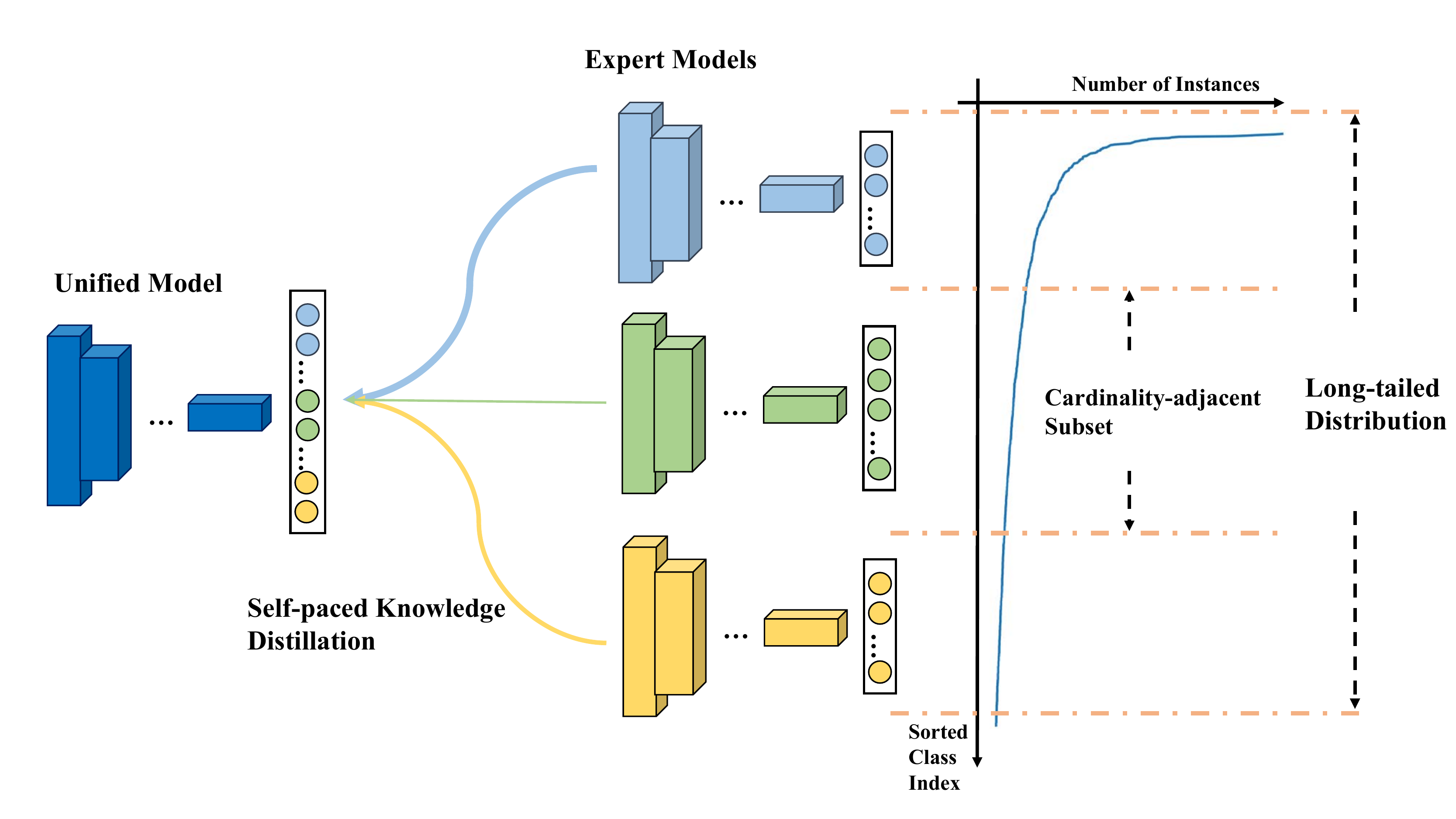}
		\caption{Schematic illustration of our proposed method.}
		\label{fig:illustration}
	\end{figure}
	
	Deep convolutional neural networks (CNNs) have achieved remarkable success in various computer vision applications such as image classification, object detection and face recognition. Training a CNN typically relies on carefully collected large-scale datasets, such as ImageNet \cite{deng2009imagenet} and MS COCO \cite{lin2014microsoft} with hundreds of examples for each class. 
	However, collecting such a uniformly distributed dataset in real-world scenarios is usually difficult since the underlying natural data distribution tends to exhibit a long-tailed property 
	with few majority classes (head) and large amount of minority classes (tail) \cite{ouyang2016factors,zhang2017range,oksuz2019imbalance}. 
	When deep models are trained under such imbalanced distribution, they are unlikely to achieve the expected performances which necessitates developing relevant algorithms. 
	
	Recent approaches tackle this problem mainly from two aspects. First is via re-sampling schemes or cost-sensitive loss functions to alleviate the negative impact of data imbalance. Second is by head-to-tail knowledge transfer, where prior knowledge or induction bias is learned from the richly annotated classes and generalize to the minority ones. 
	
	Orthogonal to the above two perspectives, we propose a novel self-paced knowledge distillation method which can be easily plugged into previous methods. Our method is motivated by an interesting observation that learning a more uniform distribution with fewer samples is sometimes easier than learning a long-tailed distribution with more samples \cite{ouyang2016factors}. We first introduce four metrics to measure the \emph{`longtailness'} of a long-tailed distribution. We then show that if we sort all the categories according to their cardinality, then splitting the entire long-tailed dataset into subsets will lead to a smaller \emph{longtailness}, which indicates that they suffer a less severe data imbalance problem. Therefore training a CNN on these subsets is expected to perform better than their jointly-trained counterparts.
	For clarity, we refer to such a subset as \textbf{cardinality-adjacent subset}, and the CNN trained on these subsets as \textbf{Expert Models}.
	
	Once we acquire the well-trained expert models,  they can be utilized as guidance to train a unified student model. If we take a look at human learning process as students, we can conclude two characteristics: 1) the student often takes various courses from easy to hard, 2) as the learning proceeds, the student acquires more knowledge from self-learning than from teachers and he/her may even exceed his/her teachers. Inspired by these findings,
	we propose a \emph{Learning From Multiple Experts (LFME)} framework with two levels of adaptive learning schemes, termed as self-paced expert selection and curriculum instance selection. Specifically, the self-paced expert selection automatically controls the impact of knowledge distillation from each expert, so that the learned student model will gradually acquire the knowledge from the experts, and finally exceed the expert. The curriculum instance selection, on the other hand, designs a curriculum for the unified model where the training samples are organized from easy to hard, so that the unified student model will receive a less challenging learning schedule, and gradually learns from easy to hard samples. 
	A schematic illustration of our LFME framework is shown in Fig.~\ref{fig:illustration}.
	
	To verify the effectiveness of our proposed framework, we conduct extensive experiments on three benchmark long-tailed classification datasets, and show that our method is able to yield superior performances compared to the state-of-the-art methods. It is worth noting that our method can be easily combined with other state-of-the-art methods and achieve further improvements. Moreover, we conduct extensive ablation studies to verify the contribution of each component. 
	
	Our contributions can be summarized as follows: 
	(1) We introduce four metrics for evaluating the \textit{`longtailness'} of a distribution and further propose a \emph{Learning From Multiple Experts} knowledge distillation framework.
	(2) We propose two levels of adaptive learning schemes, i.e. model level and instance level, to learn a unified \emph{Student} model. (3) Our proposed method achieves state-of-the-art performances on three benchmark long-tailed classification datasets, and can be easily combined with state-of-art methods for further improvements.

	\section{Related Work}
	\noindent\textbf{Long-tailed, Data-imbalanced Learning.} \quad The long-tailed learning problem has been comprehensively studied due to the prevalence of data imbalance problem \cite{he2009learning,ouyang2016factors}. Most previous methods tackle this problem using either re-sampling, re-weighting or `head-to-tail' knowledge transfer. Re-sampling methods either adopt over-sampling on tail classes \cite{chawla2002smote,han2005borderline} or use under-sampling \cite{drummond2003c4,tahir2012inverse,jeatrakul2010classification} on head classes. On the other hand, various cost-sensitive loss functions have been proposed in the literature to re-weight majority and minority instances \cite{khan2017cost,zhang2017range,dong2017class,lin2017focal,huang2019deep,cao2019learning,cui2019class}. Among them, 
	Range Loss \cite{zhang2017range} minimizes the range of each class to enhance the learning towards face recognition with long-tail while Focal Loss \cite{lin2017focal} down-weights the loss assigned to well-classified examples to deal with class imbalance in object detection.
	Label-Distribution-Aware Margin Loss (LDAM) \cite{cao2019learning}, on the other hand, encourages minority classes to have larger margins.
	
	Researchers also try to employ head-to-tail knowledge transfer for data imbalance. In  \cite{wang2016learning,wang2017learning} a transformation from minority classes to majority classes regressors/classifiers is learned progressively while in \cite{liu2019large} a meta embedding equipped with a feature memory is proposed for such knowledge transfer. 
	
	Few-shot learning methods \cite{hariharan2017low,finn2017model,gidaris2018dynamic,xiang2019incremental} also try to generalize knowledge from a richly annotated dataset to a low-shot dataset. This is often achieved by training a meta-learner from the many-shot classes and then generalize to new few-shot classes. However, different from few-shot learning algorithms, we mainly focus on learning a continuous spectrum of data distribution jointly, rather than focus solely on the few-shot classes.
	
	\noindent\textbf{Knowledge Distillation.} \quad 
	The idea of knowledge distillation was first introduced in \cite{hinton2015distilling} for the purpose of model compression where a student network is trained to mimic the behavior of a teacher network so that the knowledge is compressed to the compact student network. Then the distillation target is further extended to hidden layer features \cite{romero2014fitnets} and visual attention \cite{zagoruyko2016paying}, where attention map from the teacher model is transferred to the student. Apart from distilling for model compression, knowledge distillation is also proved to be effective when the teacher and the student have identical architecture. i.e. self-distillation \cite{yim2017gift,furlanello2018born}. Knowledge distillation is also applied in other areas such as continual learning \cite{li2017learning,rebuffi2017icarl}, semi-supervised learning \cite{lopez2015unifying} and neural style transfer \cite{huang2017like}. 
	
	\noindent\textbf{Curriculum and Self-paced Learning.} \quad
	The basic idea of curriculum learning is to organize samples or tasks in ascending order of difficulty, and it has been widely adopted for weakly supervised learning \cite{li2016weakly,jiang2017mentornet,guo2018curriculumnet} and reinforcement learning\cite{svetlik2017automatic,narvekar2017curriculum,narvekar2017autonomous}. Apart from designing easy-to-hard curriculums based on prior knowledge, efforts have also been made to incorporate learning process to dynamically adjust the curriculum. In \cite{jiang2015self} a self-paced curriculum determined by both prior knowledge and learning dynamics is proposed. In \cite{graves2017automated} a multi-armed bandit algorithm is used to determine a syllabus, where the rate of increase in prediction accuracy and network complexity are utilized as reward signals. In \cite{ren2018learning}, meta learning is employed to assign weights to training samples based on gradient directions. 

	\begin{table}
		\begin{center}
			\caption{Comparison of entire distribution and subsets under four metrics. Larger values indicate more \emph{longtailness}.}
			\begin{tabular}{l|c|c|c|c}
				\hline
				
				\diagbox{Metric}{Accuracy} & $I_{Ratio}$  & $I_{KL}$ & $I_{Abs}$ & $I_{Gini}$ \\
				\hline\hline
				Entire & 256.0 & 0.707 & 0.769 & 0.524 \\
				
				Many-shot & 12.8 & 0.278 & 0.481 & 0.322 \\
				
				Medium-shot & 4.7 & 0.122 & 0.356 & 0.235 \\
				
				Low-shot & 4.0 &  0.099 & 0.320 & 0.209 \\
				\hline
			\end{tabular}
		\end{center}
		
		\label{table:metrics}
	\end{table}

	\section{Motivation and Metrics for Evaluating Data Imbalance}
	
	The problem we address in this work is to train a CNN on a long-tailed classification task. Our method is inspired by an interesting empirical finding that, training a CNN on a balanced dataset with fewer samples sometimes leads to superior performances than on a long-tailed dataset with more samples. As the experiment in \cite{ouyang2016factors} reveals that even when 40\% of the positive samples are left out for the representation learning, the object detection performances can still be surprisingly improved a bit due to a more uniform distribution. This observation successfully emphasizes the importance of learning a balanced, uniform distribution. 
	To learn a more balanced distribution, a natural question to ask would be, how to measure the data imbalance. Since almost every manually collected dataset more or less contains various number of samples per class.
	To this end, we first introduce four metrics for data imbalance measurement.
	
	For a long-tailed dataset, if we denote $N, N_i, C$ to be the total number of samples, number of samples in class $i$, and number of classes respectively, then the four metrics for measuring data imbalance is introduced as follows:

	\noindent\textbf{Imbalance Ratio} \cite{stamatatos2008author} is defined as the ratio between the largest and the smallest number of samples: $I_{Ratio} = \frac{N_{i_{max}}}{N_{i_{min}}}  $
	
	\begin{figure}[htbp]
		\centering
		\includegraphics[width= 0.9\linewidth]{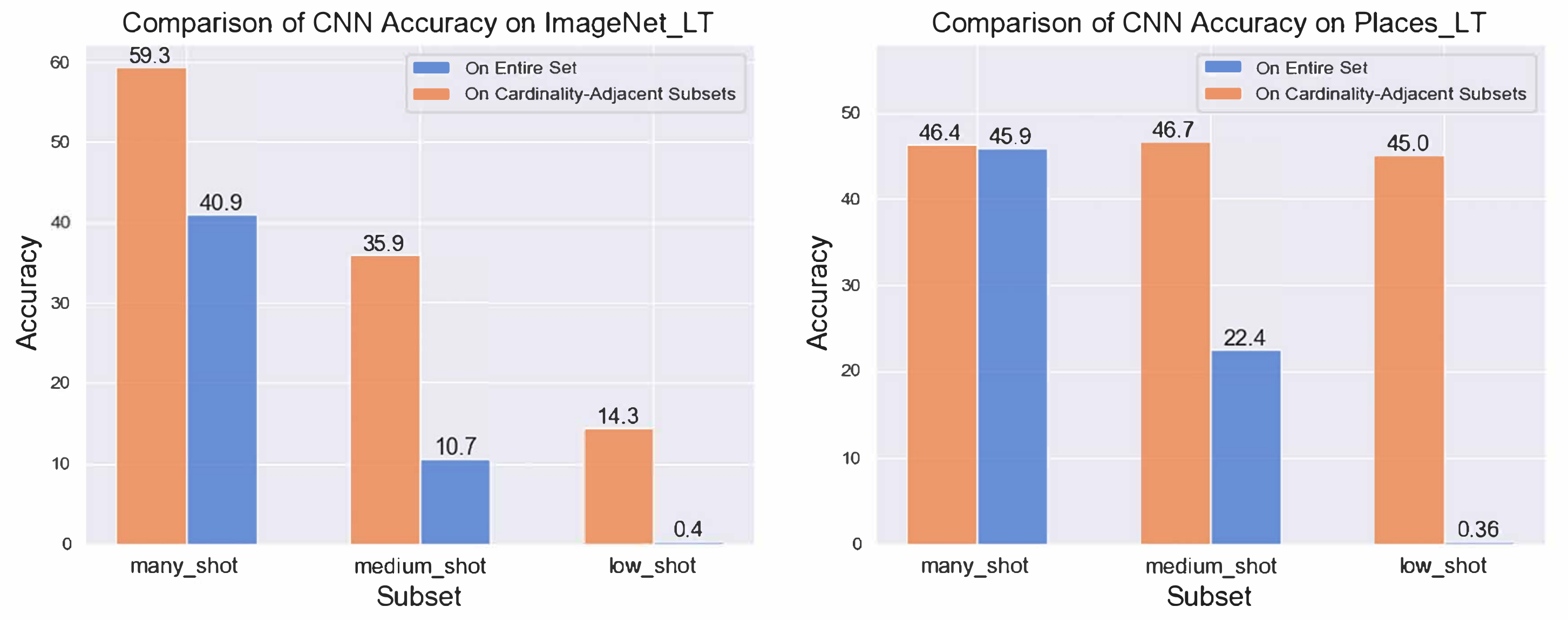}
		\caption{Comparison of CNN performances trained on cardinality-adjacent subsets and the entire dataset.}
		\label{fig:subset_vs_joint}
	\end{figure}

	\noindent\textbf{Imbalance Divergence} is defined as the KL-Divergence between the long-tailed distribution and the uniform distribution: \[I_{KL} = D(P||Q) = \sum_i p_i log \frac{p_i}{q_i}\] where $p_i=\frac{N_i}{N}$ is the class probability for class $i$, and $q_i = \frac{1}{C}$ denotes the uniform probability.
	
	\noindent\textbf{Imbalance Absolute Deviation} \cite{collins2018evolutionary} is defined as the sum of aboslute distance between each long-tailed and uniform probability: \[I_{Abs} = \sum_i |\frac{1}{C} - \frac{N_i}{N}|\]
	\noindent\textbf{Gini Coefficient} is defined as
	\[I_{Gini} = \frac{\sum_{i=1}^{C} (2i-C-1)N_i}{C\sum_{i=1}^{C}N_i}\]
	where $i$ is the class index when all classes are sorted by cardinality in ascending order.
	
	The last three metrics all measure the distance between the current long-tailed distribution and a uniform distribution. For all four metrics, smaller values indicate a more uniform distribution. Having these metrics at hand, we show that for a long-tailed dataset, if we sort the classes by their cardinality, i.e. number of samples, then a subset of the classes with adjacent cardinality will become less long-tailed under these imbalance measurements. As Table \ref{table:metrics} shows that, if we split the long-tailed benchmark dataset ImageNet-LT into three splits (many-shot, medium-shot, few-shot) according to class cardinality (following \cite{liu2019large}), then all four metrics become smaller, which indicates that these subsets become less imbalanced than the original. Then the CNNs trained on these subsets are expected to perform better than their jointly-trained counterparts. We verify this assumption on two long-tailed benchmark datasets ImageNet-LT and Places-LT \cite{liu2019large}. As shown in Fig. \ref{fig:subset_vs_joint} that CNNs trained on these subsets outperform the joint model by a large margin. This empirical result also accords with our intuition that a continuous spectrum subset of adjacent classes is more balanced in terms of cardinality, and the learning process involves less interference between the majority and the minority.

	\section{The LFME framework}
	\subsection{Overview}
	
	\begin{figure*}[htbp!]
		\centering
		\includegraphics[width= 0.9\linewidth]{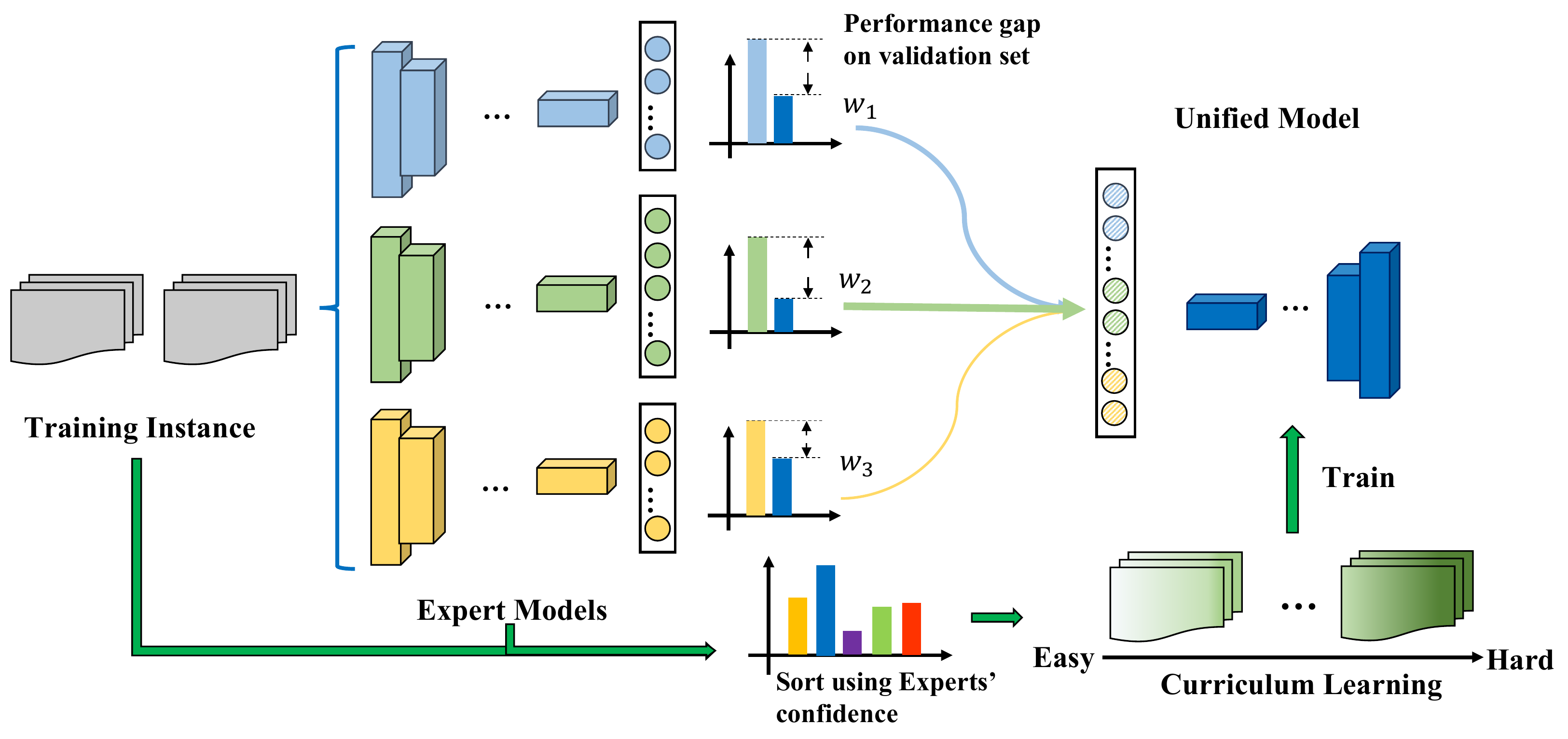}
		\caption{Overview of the LFME framework.}
		\label{fig:model}
	\end{figure*}

	In this section, we describe the proposed LFME framework in detail. Formally, given a long-tailed dataset $D$ with $C$ classes, we split the entire set of categories into $L$ cardinality-adjacent subsets $S_1, S_2,..., S_L$ using $L-1$ thresholds $T_1, T_2, ..., T_{L-1}$ such that $S_i = \left\{  c  | T_i \leq N_c \leq T_{i+1} \right\}$, where $N_c$ is class $c$'s cardinality. 
	Then we train $L$ expert models on each of the cardinality-adjacent subset and denote them as $\mathcal{M}_{E_1}, \mathcal{M}_{E_2}, ..., \mathcal{M}_{E_L}$. These expert models serve to 1) provide output logits' distribution for knowledge transfer 2) provide output confidence as instance difficulty cues. 
	These information enables us to develop self-paced and curriculum learning schemes from both model level and instance level. From the perspective of \textbf{self-paced expert selection}, we adopt a weighted knowledge distillation between the output logits from the expert models and the student model. As the learning proceeds, the student will gradually approach the experts' performances. In such cases, we do not want the experts to limit the learning process of the student. To achieve this goal, we introduce a self-paced weighted scheme based on the  performance gap on the validation set between the expert models and the student model. As the student model acquires knowledge from both data and the expert models, the importance weight of knowledge distillation will gradually decrease, and finally the unified student model is able to achieve comparable or even superior performance compared to the experts. From the perspective of \textbf{curriculum instance selection}, given the confidence scores from the Expert models, we re-organize the training data from easy to hard, i.e. from low-confidence to high-confidence. Then we exploit a soft weighted instance selection scheme to conduct such curriculum, so that easy samples are trained first, then harder samples are added to the training set gradually. This progressive learning curriculum has proved to be beneficial for training deep models \cite{bengio2009curriculum}. Finally, with the two levels of self-paced and curriculum learning schemes, the knowledge from the expert models will be gradually transferred to the unified student model.
	An overview of the LFME framework is shown in Fig \ref{fig:model}. 
	
	\subsection{Self-paced Expert Selection}
	Once we acquire the well-trained expert models, we feed the training data and obtain their output predictions. Then we employ knowledge distillation as an extra supervision signal to the student model. 
	Specifically, for the expert $\mathcal{M}_{E_l}$ trained on $l$-th cardinality-adjacent subset $S_l$, 
	if we denote $\mathbf{z}^{(l)}, \hat{\mathbf{z}}$ to be the output logits of the current expert model and the current student model respectively, then the knowledge distillation loss for expert $\mathcal{M}_{E_l}$ is given by: 
	\[L_{KD_l} = - H(\tau(\mathbf{z}^{(l)}), \tau(\hat{\mathbf{z}}^{(l)})) = - \sum_{i=1}^{|S_l|} \tau(z_i^{(l)}) \log (\tau(\hat{z_i}^{(l)}))\]
	where $\hat{\mathbf{z}}^{(l)} = \hat{\mathbf{z}}_{c \in S_l}$ is the student logits for classes in $S_l$ and 
	\[ \tau(z_i^{(l)}) = \frac{\exp(z_i^{(l)}/T)}{\sum_j \exp(z_j^{(l)}/T)}, \quad \tau(\hat{z_i}^{(l)}) = \frac{\exp(\hat{z_i}^{(l)}/T)}{\sum_j \exp(\hat{z_j}^{(l)}/T)}  \]
	are the output probabilities using Softmax with temperature  $T$. $T$ is usually set to be greater than 1 to increase the weight for smaller probabilities.
	In this way, for each expert model $\mathcal{M}_{E_l}$ we have its knowledge distillation loss to transfer its knowledge to the student model, and there are $L$ losses in total, corresponding to $L$ experts trained on $L$ cardinality-adjacent subsets. The most straightforward way to aggregate these losses would be simply summing them up. However, this could be problematic, since as the learning process goes on, the student model's performance will gradually approach or even exceed the expert's. In such cases, we do not want the expert models become performance ceilings for the student model, and we wish to gradually weaken the guidance from the experts as the data-driven learning proceeds.
	
	To achieve this goal, we propose a Self-paced Expert Selection scheme based on the performance gap between the student and the experts. In the experiments, we use the Top-1 Accuracy on the validation set after each training epoch as the measurement for performance gap. If we denote the Top-1 Accuracy of the student model $\mathcal{M}$ and the expert model $\mathcal{M}_{E_l}$ at epoch $e$ to be $Acc_{\mathcal{M}}$ and $Acc_{E_l}$ respectively, then a weighting scheme is defined as follows:
	
	$$ w_l=\left\{
	\begin{aligned}
	& 1.0 \quad & if Acc_{\mathcal{M}} \leq \alpha Acc_{E_l}  \\
	& \frac{Acc_{E_l} - Acc_{\mathcal{M}}}{Acc_{E_l}(1-\alpha)} & if Acc_{\mathcal{M}} > \alpha Acc_{E_l} 
	\end{aligned}
	\right.
	$$
	and $w_l$ is updated at the end of each epoch. The weight scheduling threshold $\alpha$ controls the knowledge distillation to switch from ordinary to a self-paced decaying schedule.
	With the self-paced weight scheduling weight $w_l$, the knowledge transfer from the experts is automatically controlled by the student model's performance. 
	The final knowledge distillation loss is the automatic weighted sum of knowledge distillation loss from all expert models:
	\[L_{KD} = \sum_{l=1}^{L} w_l L_{KD_l}\]
	
	\subsection{Curriculum Instance Selection}
	Following the spirit of curriculum learning which mimics the human learning process, three questions need to be answered: 1) how to evaluate the difficulty of each instance, 2) how to \emph{select} or \emph{unselect} a sample, 3) how to design a curriculum so that samples are organized from easy to hard.
	
	For the first question, we use the expert models' output confidence for each instance as an indication for instance difficulty. Given a training instance $(x_i, y_i)$, suppose its ground-truth class $y_i$ falls into the $l$-th subset $S_l$, i.e. $y_i \in S_l$, then we take the corresponding $l$-th expert model and use its output prediction for class $y_i$ as confidence score, denoted as $p_i$. In this way, we can obtain the confidence score for all the instances in the training set.

	For the second question, we adopt a soft selection method for instance selection. For instance $(x_i, y_i)$, we replace the cross entropy loss with a weighted version:
	\[ L_{CE} = \sum_{i=1}^N v_i^{(k)} L_{CE}(x_i)\]
	where $v_i^{(k)} \in [0, 1]$ is the selection weight at $k$-th epoch. A higher value of $v_i$ (close to 1) indicates a soft selection of $i$-th instance, while a smaller value indicates a soft unselection of that instance. 
	
	Finally, to answer the third question, we design an automatic curriculum to determine the value of $v_i^{(k)}$, so that the instances are selected from easy to hard.
	The simplest approach is to sort the instances using their confidence score $p_i$ obtained by the expert models. However, different from traditional curriculum learning scenarios, our long-tailed classification problem involves both many-shot and low-shot categories, where low-shot instances tend to have lower confidence scores than many-shot instances. When sorted by the confidence score, the low-shot samples tend to be classified as hard examples and are not selected at first, which we do not wish to happen. To deal with such scenarios, instead of sorting across the whole training set, we sort instances according to their confidence scores \textbf{within each cardinality subset}. To be more specific, given the expert output confidence, $v_i^{k}$ should be determined by three factors 1) the expert confidence $p_i$, 2) current epoch $k$,  3) the cardinality-adjacent subset $S_l$ the $i$-th instance belongs to. Since the whole dataset is long-tailed, while we select samples from easy to hard, we also wish to select as uniform as possible across all subsets at the beginning of the training, and gradually add more hard samples as the epoch increases. In other words, at the first epoch we wish to select all the samples in the subset with lowest shots $\mathcal{S}_{min}$ (i.e. classes in $\mathcal{S}_{min}$ have the smallest number of samples) and same amount of samples in other subsets, and gradually add more samples until all the samples in all subsets are selected in the last epoch.
	
\begin{figure}[htbp!]
		\centering
		\includegraphics[width=0.8\linewidth]{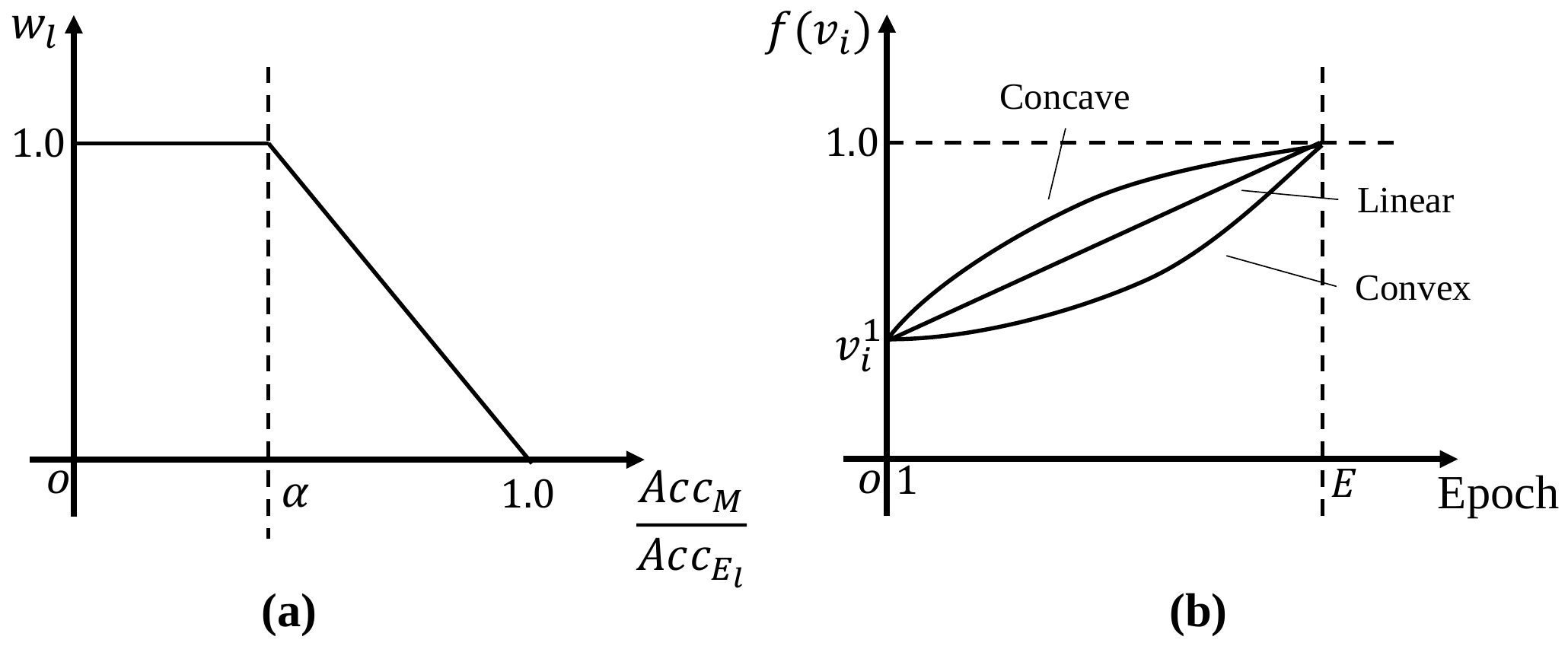}
		\caption{Weight scheduling function for (a) model level selection $w_l$, and (b) instance level selection $v_i$.}
		\label{fig:schedule}
	\end{figure}

	To achieve this goal, if we denote $N_{\mathcal{S}_l} = \frac{1}{|\mathcal{S}_l|} \sum_{i=1}^{|\mathcal{S}_l|} N_i$ as the average shot (average number of samples per class) in subset $\mathcal{S}_l$, then $v_i^{(k)}$ is determined by $p_i\frac{N_{\mathcal{S}_{min}}}{N_{\mathcal{S}_{l}}}$ at epoch 1, and grows gradually to 1 at the last epoch. Then at epoch 1, each subset softly selects its  $N_{\mathcal{S}_{min}}$ easiest samples, and harder samples are gradually softly added to the training process. 
	Formally, we use a monotonically increasing function $f$ as scheduling function, so that $ v_i^{(k)} $ will gradually grow from $p_i\frac{N_{\mathcal{S}_{min}}}{N_{\mathcal{S}_{l}}}$ to 1. For simplicity, we choose the linear function in the experiments and $f$ is defined as 
	\[ f(v_i^{k}) = (1-v_{i}^{(1)})\frac{e}{E}+v_{i}^{(1)}\]
	where $v_{i}^{(1)} = p_i\frac{N_{\mathcal{S}_{min}}}{N_{\mathcal{S}_{l}}}$ is the initial soft selection weight at epoch 1, and $e, E$ are the current epoch and the total number of epochs respectively. It is worth noting that the scheduling function $f$ can also be any convex or concave function as long as it is monotonically increasing within $[1,E]$. The impact of choosing different $f$ is further analyzed in the experimental section. A schematic illustration of $w_l$ and $v_i$ can be found in Fig \ref{fig:schedule}.
	
	\subsection{Training}
	With the Self-paced Expert Selection and Curriculum Instance Selection, we obtain the final loss function:
	\[L = \sum_{i=1}^{N} v_i L_{CE}(x_i, y_i) + \sum_{l=1}^{L}\sum_{i=1}^N w_l L_{KD_l}(\mathcal{M}, \mathcal{M}_{Exp}; x_i)\]
	where $N, L$ are the number of training instances and number of experts respectively, and $v_i, w_i$ controls the two levels of adaptive learning schedules. In practice, we first train expert models using ordinary Instance-level Random  Sampling, where each instance  is  sampled with equal probability. We then train the whole LFME using Class-level Random Sampling adopted in \cite{liu2019large,kang2019decoupling}, where each class is sampled with equal number of samples and probability.



	\section{Experiments}
	
	\subsection{Experimental Settings}
	\noindent\textbf{Dataset} \quad We evaluate our proposed method on three benchmark long-tailed classification datasets: ImageNet-LT, Places-LT proposed in \cite{liu2019large} and CIFAR100-LT proposed in \cite{cao2019learning}. ImageNet-LT is created by sampling a subset of ImageNet \cite{deng2009imagenet} following the Pareto distribution with power value $\alpha=6$. It contains 1000 categories with class cardinality ranging from 5 to 1280. Places-LT is created similarly from Places dataset \cite{zhou2017places} and contains 365 categories with class cardinality ranging from 5 to 4980. CIFAR100-LT is created with exponential decay imbalance and controllable imbalance ratio.
	
	\noindent\textbf{Baselines} \quad
	For the first two datasets, similar to \cite{liu2019large}, our baseline methods include three re-weighting methods: Lifted Loss \cite{oh2016deep}, Focal Loss \cite{lin2017focal} and Range Loss \cite{zhang2017range}, and one SOTA few-shot learning method FSLwF \cite{gidaris2018dynamic}, as well as the recent SOTA method OLTR \cite{liu2019large}. For the CIFAR100-LT dataset, we mainly compare with the SOTA method LDAM proposed in \cite{cao2019learning}.

	\noindent\textbf{Implementation Details} \quad
	For the first two datasets, we choose the number of cardinality-adjacent subsets $L=3$ with thresholds $\{20, 100\}$ following the splits in \cite{liu2019large}. We refer to these subsets as many, medium and few-shot subsets. For CIFAR100-LT, We equally split the 100 classes into two subsets: many and few-shot.
	We use PyTorch \cite{paszke2017automatic} to implement all experiments. For the first two datasets, we first train the experts using SGD for 90 epochs. Then the LFME model is trained with SGD with momentum 0.9, weight decay 0.0005 for 90 epochs, batch size 256, learning rate 0.1 and divide by 0.1 every 40 epochs. We use ResNet-10 \cite{he2016deep} training from scratch for ImageNet-LT and ImageNet pretrained Resnet-152 for Places-LT. During training, class-balanced sampling is adopted. For the CIFAR100-LT experiments, we first train the experts for 200 epochs and then train the LFME model using SGD with momentum 0.9, weight decay $2 \times 10^{-4}$, batch size 128, epochs 200, initial learning rate 0.1 and decay by 0.01 at 160 and 180 epochs, as well as deferred class-balanced sampling, same as \cite{cao2019learning}. The backbone network is ResNet-32.
	The distillation temperature $T$ is set to 2, and the expert weight scheduling threshold $\alpha$ is set to 0.6 during the experiments.  
	
	\begin{table}
		\begin{center}
			
			\caption{Long-tailed classification results on ImageNet-LT and Place-LT. * denotes reproduced results, other results are from \cite{liu2019large}.}
			\label{table:main}
			\begin{tabular}{l|p{1.0cm}<{\centering} p{1.0cm}<{\centering} p{1.0cm}<{\centering} p{1.0cm}<{\centering} |p{1.0cm}<{\centering} p{1.0cm}<{\centering} p{1.0cm}<{\centering} p{1.0cm}<{\centering}}
				\hline
				
				\multirow{2}*{\diagbox[width=3.0cm]{Method}{Acc.}} &
				\multicolumn{4}{c|}{ImageNet-LT } & \multicolumn{4}{c}{Places-LT } \\

				\cline{2-9} &  Many & Med. & Few & All & Many & Med.& Few & All \\
				\hline\hline
				Plain Model & 40.9 & 10.7 & 0.4 & 20.9 & \textbf{45.9} & 22.4 & 0.36 & 27.2\\
				
				Lifted Loss\cite{oh2016deep} & 35.8 & 30.4 & 17.9 & 30.8 & 41.1 & 35.4 & 24 & 35.2\\
				
				Focal Loss\cite{lin2017focal} & 36.4 & 29.9 & 16.0 & 30.5 & 41.1 & 34.8 & 22.4 & 34.6 \\
				
				Range Loss\cite{zhang2017range} & 35.8 & 30.3 & 17.6 & 30.7 & 41.1 & 35.4 & 23.2 & 35.1 \\
				
				FSLwF \cite{gidaris2018dynamic} & 40.9 & 22.1 & 15.0 & 28.4 & 43.9 & 29.9 & \textbf{29.5} & 34.9\\
				
				OLTR \cite{liu2019large} & 43.2 & 35.1 & 18.5 & 35.6 & 44.7 & 37.0 & 25.3 & 35.9 \\
				OLTR \cite{liu2019large}* & 40.7 & 33.2 & 17.4 & 33.8 & 42.2 & 38.1 & 17.8 & 35.3 \\
				\hline
				Ours  & \textbf{47.1} & 35.0 & 17.5 & 37.2 & 38.4 & 39.1 & 21.7 & 35.2 \\
				
				Ours + Focal Loss &  46.7 & 35.8 & 17.3 & 37.3 & 37.0  & \textbf{39.6} & 23.0 & 35.2 \\
				
				Ours + OLTR & 47.0 & \textbf{37.9} & \textbf{19.2} & \textbf{38.8} & 39.3 & \textbf{39.6} & 24.2  & \textbf{36.2} \\
				
				
				\hline
			\end{tabular}
		\end{center}
		
	\end{table}


	\subsection{Main Results on Long-tailed Classification Benchmarks}
	Table \ref{table:main} shows the long-tailed classification results on ImageNet-LT and Places-LT dataset. As can be found that our method is able to achieve superior or at least comparable results to the state-of-the-art methods such as OLTR. We found that  many-shot categories benefit most from our LFME framework, while few-shot classes also demonstrate improvements and perform similarly with the re-weighting methods. Moreover, we also demonstrate that our method can be easily incorporated with other state-of-the-art methods, and we show the result of LFME+Focal Loss and LFME+OLTR (where LFME is added in the second stage of OLTR). We observe that both methods benefit from our expert model on all three subsets, and the combination of our method and OLTR outperforms previous methods by a large margin. 
	It is also worth noting that our expert models are trained using vanilla CNNs, and utilizing other techniques will further lead to superior expert models, and assumably, superior student model.
	
	To further demonstrate the statistical significance of the proposed method, we conduct experiments on CIFAR100-LT with imbalance ratio 100. The results in Table \ref{table:CIFAR} show that LFME is able to achieve comparable performances with the SOTA method LDAM \cite{cao2019learning} and combining them will further improve LDAM on both many and few-shot subsets.

	
	
	
	
	
	
	

	

	\begin{table}
		
		
		
		
		
		
		
		
		
		\begin{minipage}[t]{0.50\linewidth}
			\centering
			\caption{Results on CIFAR100-LT.}
			\label{table:CIFAR}
			\begin{tabular}{l|c|c|c}
				\hline
				Methods &  Many & Few & All  \\
				\hline\hline
				Plain CNN & 59.0 & 18.2 & 38.6 \\
				Ours & 59.0 & 25.5 & 42.3  \\
				LDAM \cite{cao2019learning} & 58.8 & 26.1 & 42.4 \\
				Ours+LDAM & \textbf{59.5} & \textbf{28.0} & \textbf{43.8} \\
				\hline
			\end{tabular}
			
			
			
			
			
			
			
			
			
		\end{minipage}
		\begin{minipage}[t]{0.50\linewidth}  
			\centering
			\caption{Effect of different scheduling functions.}
			\label{table:scheduleType}
			\begin{tabular}{c|c|c|c|c}
				\hline
				Schedule &  Many & Medium & Few & All  \\
				\hline\hline
				Linear & 47.1 & \textbf{35.0} & \textbf{17.5} & \textbf{37.2} \\
				Convex & 47.2 & 34.6 & 16.7 & 36.9 \\
				Concave & \textbf{47.5} & 34.7 & 17.0 & 37.1 \\
				\hline
				
			\end{tabular}
			
		\end{minipage}
	\end{table}
	
	\subsection{Ablation Study}

	\noindent\textbf{Effectiveness of Each Component} \quad

	We evaluate each part of our method and the result is shown in Table \ref{table:ablation}. We compare with the following variants: 1) Instance-level Random Sampling (Ins.Samp.), where each instance is sampled with equal probability. 2)  Instance-level Random Sampling + Ordinary Knowledge distillation (Ins.Samp.+KD), where non-self-paced version knowledge distillation from the experts is added, i.e. $w_l=1.0$. 3) Class-level Random Sampling (Cls.Samp.), where each class is sampled with equal number of samples and probability. 4) Class-level Random Sampling + Ordinary Knowledge Distillation (Cls.Samp.+KD). 5) Class-level Random Sampling + Knowledge Distillation + Self-paced Expert Selection (Cls.Samp.+ KD + SpES). 6) Curriculum Instance Selection + Self-paced Expert Selection (CurIS+KD+SpES), which constitute our LFME framework. 
	
	\begin{wraptable}{r}{7cm}
		\caption{Effectiveness of each component.}
		\label{table:ablation}
		\begin{tabular}{l|c|c|c|c}
			\hline
			\diagbox{Method}{Accuracy} &  Many & Med. & Few & All  \\
			\hline\hline
			Ins.Samp. & 40.9 & 10.7 & 0.4 & 20.9 \\
			
			Ins.Samp.+KD &  55.7 & 22.2 & 0.02 & 32.2 \\
			
			
			Cls.Samp. & 38.8 & 32.3 & 17.0 & 32.6 \\
			
			Cls.Samp.+KD & 44.9 & 34.5 & 15.8 & 35.8 \\
			
			Cls.Samp.+KD+SpES & 46.6 & 35.8 & 16.5 & 37.1 \\
			
			CurIS+KD+SpES & 47.1 & 35.0 & 17.5 & 37.2 \\
			\hline
		\end{tabular}
	\end{wraptable}
	From the results, we come up with the following observations: first, compared to the instance-random sampling, the adopted class-level random sampling is able to largely improve the few-shot performance while also decrease the many-shot performance slightly, since it samples more few-shot and less many-shot instances. Second, the introduction of knowledge distillation from experts can significantly improve the results, as it brings \textbf{11.3\%} and \textbf{3.2\%} to the Instance-level Sampling and Class-level Sampling baselines. However, while knowledge distillation brings improvements for many-shot classes, it will also decrease the few-shot accuracy slightly.  Third, the Self-paced Expert Selection improves the knowledge distillation on all three subsets. It removes the performance ceiling brought from the experts and allows the student to exceed the experts. As the results show that SpES brings \textbf{0.4\%} and \textbf{1.3\%} overall performance gain respectively. Finally, the proposed Curriculum Instance Selection further improves on the few-shot categories with \textbf{1.0\%} in accuracy, so that the decrease on few-shot subset caused by the knowledge distillation is compensated.


	\begin{figure}[htbp!]
		\centering
		\includegraphics[width=0.8\linewidth]{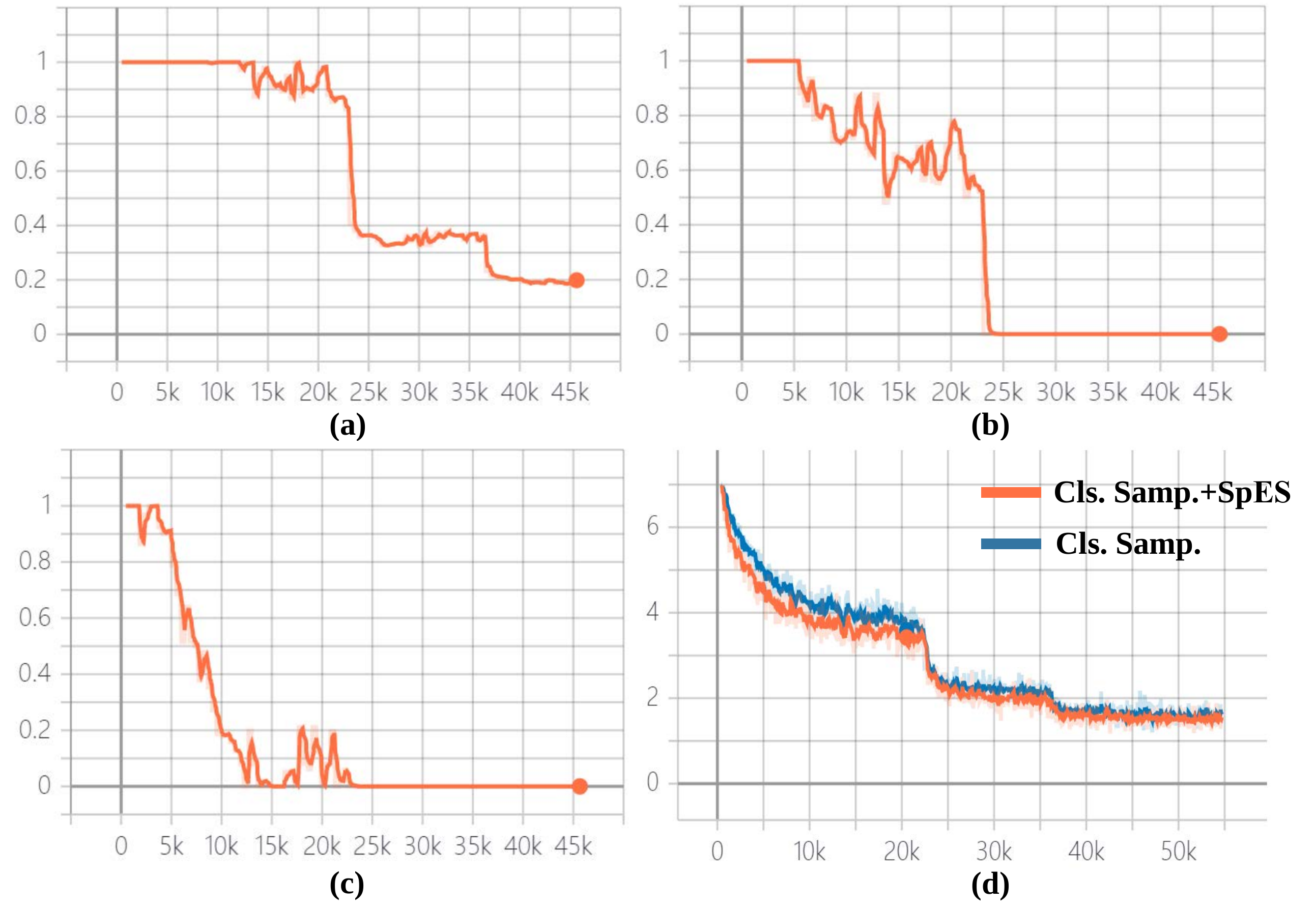}
		\caption{(a)-(c): Visualization of self-paced expert selection scheduler $w_l$ for many-shot, medium-shot, few-shot expert model. (d): Loss curves before and after adding Self-paced Expert Selection.}
		\label{fig:SpES}
	\end{figure}

	\noindent\textbf{Visualization of Self-paced Expert Selection} \quad
	Self-paced expert selection plays an important role in LFME for more efficient and effective knowledge transfer. Fig \ref{fig:SpES} (a)-(c) gives a visualization of the expert selection weights $w_l$ for many-shot, medium-shot, few-shot model. From the visualization, we observe that $w_l$ serves to automatically control the knowledge transfer, as for many-shot and medium-shot experts the knowledge is consistently distilled, while for few-shot experts, the student instantly exceeds the expert's performance, thus leading to a decay in $w_{fewshot}$. Moreover, we also visualize the impact of Self-paced Expert Selection in terms of cross-entropy loss curves, shown in Fig \ref{fig:SpES} (d), and we find that it leads to a lower cross-entropy loss, which also verifies the effectiveness of the proposed Self-paced Expert Selection.


	
	\begin{figure}[htbp!]
		\centering
		\includegraphics[width=0.8\linewidth]{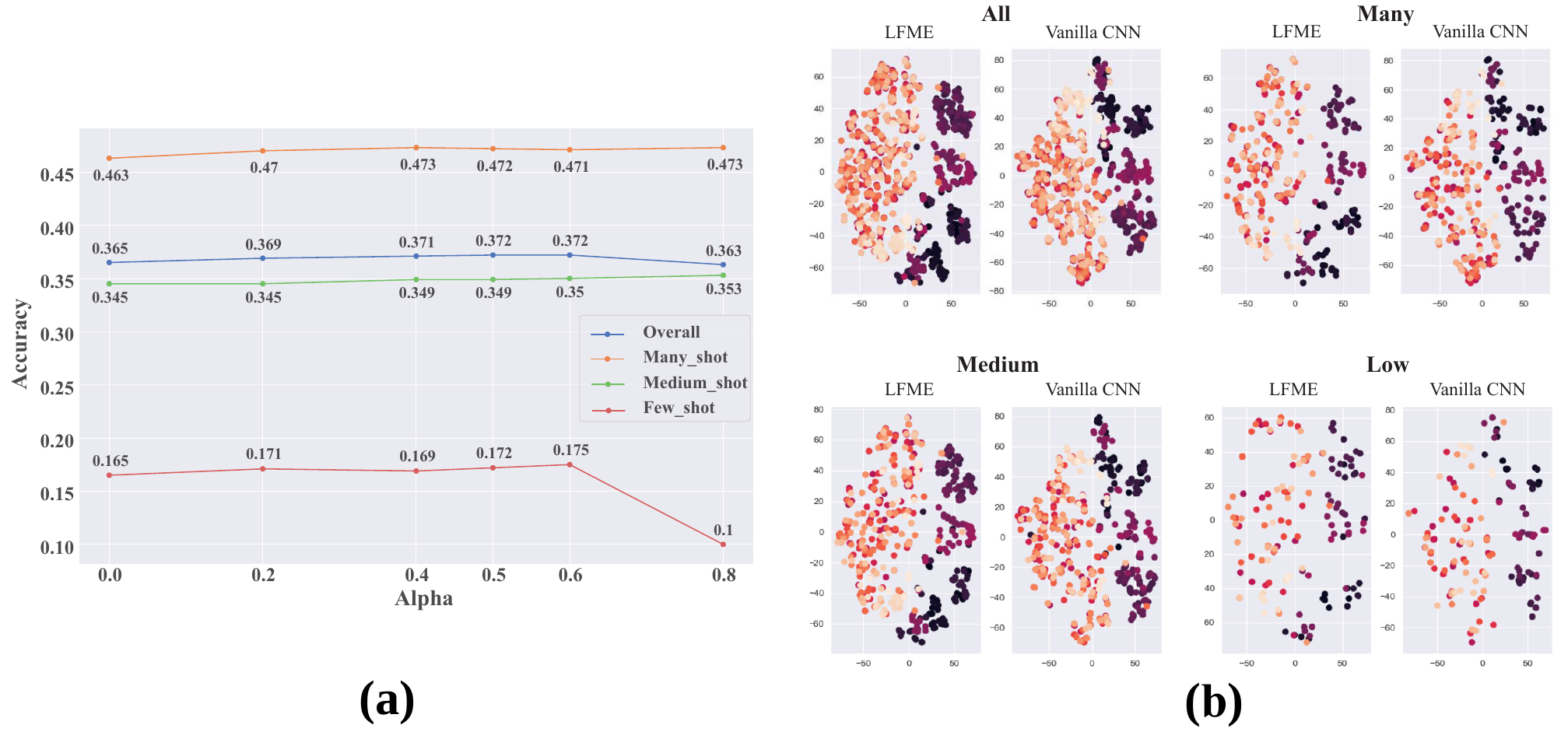}
		\caption{(a) Sensitivity analysis of $\alpha$. (b) T-SNE visualization of classification weights.}
		\label{fig:tsne_sensitivity}
	\end{figure}
	
	\noindent\textbf{Effect of Learning Scheduler} \quad
	We also discuss the impact of different learning schedules for $v_i$ as shown in Table \ref{table:scheduleType}. Given the initial instance confidence $v^{(1)}$, we test with the following scheduling functions: 
	\begin{itemize}
		\item $f_{Linear} = (1-v^{(1)})\frac{e}{E} + v^{(1)}$
		\item $f_{Convex} = 1 - (1-v^{(1)}) \cos(\frac{e}{E}\frac{\pi}{2})$
		\item $f_{Concave} = (1-v^{(1)}) \frac{\log(1+\frac{e}{E})}{\log2}$
	\end{itemize}
	The result shows that the linear growing function yields the best result, while the concave the convex function $f$ also produce similar performances. The convex function yields the worst performances as it selects fewest instances at the start of the training which may not be beneficial for the training dynamics.

	\noindent\textbf{Sensitivity Analysis of Hyperparameter $\alpha$} \quad
	Fig. \ref{fig:tsne_sensitivity}(a) shows the sensitivity analysis of expert weight scheduling threshold $\alpha$. From the result, we observe that our model is robust to most $\alpha$ values. When $\alpha$ grows to 1.0, the Self-paced Expert Selection becomes ordinary knowledge distillation, and result in a performance decline.

	\noindent\textbf{Visualization of Classification Weights} \quad
	We visualize the classification weights
	of vanilla CNN and our LFME via T-SNE in Fig  \ref{fig:tsne_sensitivity}(b). The results show that our method results in a more structured, compact feature manifold. It shows that without particular re-weighting, our method is also able to produce discriminative feature space and classifiers.

	


	
	\section{Conclusions}
	In this paper, we propose a Learning From Multiple Experts framework for long-tailed classification problem. By introducing the idea of cardinality-adjacent subset which is less long-tailed, we train several expert models and propose two levels of adaptive learning to distill the knowledge from the expert models to a unified student model. From the extensive experiments and visualizations, we verify the effectiveness of our proposed method as well as each of its component, and show that the LFME framework is able to achieve state-of-the-art performances on the long-tailed classification benchmarks.

	\section*{Acknowledgement}
	This work was supported by the National Natural Science Foundation of China (No.U1936202, 61925107). We also thank anonymous reviewers for their constructive comments. Corresponding author: Guiguang Ding.
	

	%
	%
	\bibliographystyle{splncs04}
	\bibliography{egbib}
\end{document}